\documentclass{article}  
\usepackage[margin=1.1in]{geometry}
\usepackage{lipsum}                     
\usepackage{xargs}                      
\usepackage[pdftex,dvipsnames]{xcolor}  
\usepackage{graphicx}
\usepackage[autostyle]{csquotes}
\usepackage{subcaption}
\usepackage[]{amsmath}
\let\labelindent\relax
\usepackage[inline]{enumitem}
\usepackage{cite}

\usepackage[colorinlistoftodos,prependcaption,textsize=tiny]{todonotes}
\newcommandx{\unsure}[2][1=]{\todo[linecolor=red,backgroundcolor=red!25,bordercolor=red,#1]{#2}}
\newcommandx{\change}[2][1=]{\todo[linecolor=blue,backgroundcolor=blue!25,bordercolor=blue,#1]{#2}}
\newcommandx{\info}[2][1=]{\todo[linecolor=OliveGreen,backgroundcolor=OliveGreen!25,bordercolor=OliveGreen,#1]{#2}}
\newcommandx{\improvement}[2][1=]{\todo[linecolor=Plum,backgroundcolor=Plum!25,bordercolor=Plum,#1]{#2}}
\newcommandx{\thiswillnotshow}[2][1=]{\todo[disable,#1]{#2}}

\title{\LARGE \bf Robot Communication Via Motion:\\ \mbox{Closing the Underwater Human-Robot Interaction Loop}\footnote{This work is under review for ICRA 2019.}}

\author{Michael Fulton$^{1}$, Chelsey Edge$^{2}$, Junaed Sattar$^{3}$
\thanks{The authors are from the University of Minnesota, Twin Cities, Minneapolis, MN, USA.
        {\tt\small \{$^{1}$fulto081, $^{2}$edge0037, $^{3}$junaed\}@umn.edu}}%
}

\begin{document}

\maketitle
\pagestyle{plain}

\begin{abstract}
In this paper, we propose a novel method for underwater robot-to-human communication using the motion of the robot as \enquote{body language}.  
To evaluate this system, we develop simulated examples of the system's body language gestures, called kinemes, and compare them to a baseline system using flashing colored lights through a user study. 
Our work shows evidence that motion can be used as a successful communication vector which is accurate, easy to learn, and quick enough to be used, all without requiring any additional hardware to be added to our platform. 
We thus contribute to \enquote{closing the loop} for human-robot interaction underwater by proposing and testing this system, suggesting a library of possible body language gestures for underwater robots, and offering insight on the design of nonverbal robot-to-human communication methods.
\end{abstract}
\section{INTRODUCTION}
In the United States of America as of May 2017, 3,280 people are employed as commercial divers, tasked with dangerous and difficult tasks underwater~\cite{commercial_nodate}. The presence of a capable Autonomous Underwater Vehicle (AUV) as a partner to assist in data collection, monitoring, search and rescue, or maintenance tasks has the potential to increase efficiency and ensure safety of the diver~\cite{sattar_enabling_2008}. In order for an AUV to be an effective partner to a human, it must be capable of accurate and efficient communication with its partner. A number of methods have been proposed to enable humans to communicate with robots underwater~\cite{sattar_fourier_2007,islam_dynamic_2017,islam_follow_2018}, but few have addressed the inverse problem of how the robot could communicate back.
Underwater, the well-explored interaction vectors of voice interaction (through speech synthesis and text-to-speech systems) and text interaction (through keyboards and screens) are infeasible or less effective. It is therefore desirable to develop new modalities of \textit{robot-to-human communication} for underwater robots to enable their use as workers, companions, and guides to divers. 

Robot communication to humans underwater is quite challenging, as the two most common modalities used for human interaction are significantly limited. Sound is distorted and attenuated and can be masked by equipment noise. While vision is usually available, its quality is often degraded\cite{fabbri_ehancing_2018}\cite{lu_underwaterimage_2016}. In such a challenging environment, the de facto solution is to simply accept the AUV as a silent partner. This is an issue, as it keeps AUVs from achieving their full potential as diver partners by offering relevant safety information, providing advice, and most importantly, engaging in dialogue with humans.
A common method for underwater robot-to-human communication is the use of a display device such as an OLED or LCD display, integrated into the robot, as in the Aqua AUV~\cite{dudek_aqua_2007}. However, such displays are typically very small and hard to read from a distance or at an angle, introduce additional weight and power requirements, and are susceptible to failure. The other primary method that is proposed and has been used is messaging via flashing lights~\cite{demarco_underwater_2014}. This has the advantage of wider viewing angles and the fact that many AUVs already have some kind of lighting system. However, lights are not a natural communication vector and require divers to commit to memory a list of light codes and associated meanings. They also may have a limited number of possible meanings which can be communicated, since most built-in systems have only one light and are thus limited to varying the blinking rate of that single light.

\begin{figure}
\centering
\includegraphics[width=0.8\linewidth]{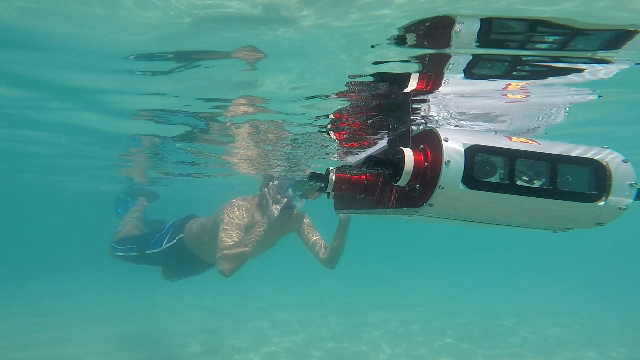}
\label{fig:aqua_interact}
\caption{An example of a diver interacting with the Aqua AUV, instructing the robot with free-form hand gestures.}
\end{figure}


In response to the drawbacks of the systems described above, we construct a list of desiderata.
Underwater robot-to-human communication should:
\begin{enumerate*}[label=\textbf{\alph*})]
	\item work from a distance and multiple viewing angles,
	\item require no additional hardware,
	\item be natural and easy to learn, and
	\item allow for a large number of interaction phrases.
\end{enumerate*}
To address these desiderata, we propose the use of robot motion as \textbf{kinemes}~\cite{nöth_handbook_of_semiotics_1995}, a motion associated with a distinct meaning. We believe that a kineme communication system fulfills our desiderata, while remaining fast enough to be feasible for underwater interaction tasks. 

To validate the use of motion as a communication technique, we conducted a study using simulated videos of the Aqua AUV to test the accuracy, efficiency, and ease of learning provided by such a system. Twenty-four participants tested the system against a baseline system comprised of colored lights flashing in codes, and the resultant data was analyzed to determine whether motion-based communication could be adequately accurate, efficient, and easy to learn for use in underwater robot-to-human communication. 

In this paper, we make the following contributions:

\begin{itemize}
	\item We propose a unique system: motion-based (kineme) communication for underwater robots which purely uses motion to communicate information to human collaborators.
    \item We show that there is a statistically significant improvement in accuracy of communication when using kinemes compared to light codes.
    \item We show that kinemes outperform light codes in ease of learning and that given enough education, they can be nearly as quick to understand as light codes.  
\end{itemize}

We also provide insight into the design of motion-based communication systems for other 6-DOF robots (robots which translate and rotate freely in three dimensional space), such as which kinds of information are best suited for kineme communication and discuss potential design implications for the implementation of our system using the Aqua AUV. 
\section{RELATED WORK}

\subsection{Underwater Human Robot Interaction}


\subsubsection{Diver Communication}
Underwater human-robot interaction (HRI) primarily focuses on the problem of human-to-robot communication. In this space, the emphasis has been on removing barriers to communication between the pair. A common approach is for a human operator on the surface to interpret the needs of the diver (possibly through hand signals from the diver) and teleoperate the AUV accordingly. This could be considered \textit{direct control}, supported by high-speed tethered communication~\cite{aoki_transmission_1997}. However, having the controller on the surface introduces latency and causes some issues by limiting situational awareness. In order to remove this latency, the use of waterproofed control devices at depth has been proposed \cite{verzijlenberg_tablet_2010}, which essentially moves the controller down to the depth of the robot. This method requires additional hardware however, which adds weight and complexity to the interaction scenario. To avoid this, onboard systems to enable human-to-robot communication are preferable.
Examples of systems of communication which do not require an additional device include the use of fiducial tags~\cite{sattar_fourier_2007}\cite{dudek_visual_language_2007} or of hand gestures~\cite{islam_dynamic_2017}\cite{demarco_underwater_2014}, which can be recognized and interpreted onboard a robot.

\subsubsection{Robot Communication}
As previously mentioned, the inverse communication problem of the robot communicating to the human has not been well explored. In many methods, the robot responds to the human's input via a small display ~\cite{sattar_enabling_2008, sattar_ensuring_2014,islam_dynamic_2017}. These displays are typically difficult to read from any distance or outside of optimal viewing angles. Larger screens, as in \cite{ukai_swimoid:_2013}, would greatly reduce the depth rating and effective range of an AUV.
Specific interaction devices generally are used for bidirectional communication, as in \cite{verzijlenberg_tablet_2010}. Once again, this requires the addition of other devices, which is costly and increases complexity. One of the more unique proposals is the use of an AUV's light system in \cite{demarco_underwater_2014}, where illumination intensity is modulated to communicate simple ideas. This case study proved that a robot and a human could collaborate to achieve a task underwater, but the communication methods used were not validated by a multi-user study.


\subsection{Nonverbal, Non-facial and Non-Humanoid HRI}
Nonverbal methods form only a small portion of HRI, much of which focuses on displaying emotions in humanoid platforms, therefore, their results are only  tangentially related to the problem of nonverbal underwater HRI, as our focus is non-humanoid robots displaying information rather than emotion. That said, there are a number of works which directly relate to our problem of robot-to-human communication using nonverbal, non-facial methods with a non-humanoid robot. These can be classified into categories based on their intended purpose. A useful source for these categories is the work of Mark L. Knapp \cite{knapp1972nonverbal}, which defines five basic categories of body language:

\begin{enumerate}
	\item Emblems, which have a particular linguistic meaning.
    \item Illustrators, which provide emphasis to speech.
    \item Affective display, which represent emotional states.
    \item Regulators, which control conversation flow.
    \item Adapters, which convey non-dialogue information.
\end{enumerate}

The bulk of nonverbal, non-facial, and non-humanoid HRI is concerned with affective display, while our work is primarily concerned with emblems.

\subsubsection{Emblems}
Emblems in nonverbal, non-facial HRI on non-humanoid platforms should be body movements which code directly to some linguistic meaning. This is the area to which our kineme communication system belongs, though it has little company here. 
The previously mentioned case study by DeMarco et al. \cite{demarco_underwater_2014} is one of the few attempts to communicate information rather than emotion using non-verbal methods, in this case via changes in the illumination of a light. Another example of emblems in this type of communication is the up-and-down tilt of a pan-tilt camera being used as a nod in~\cite{wainer_role_2006}, in which the robot simulates head gestures by controlling its camera's pan and tilt.

\subsubsection{Affective Display}

Affective display is by far the most explored realm of non-verbal communication with non-humanoid robots.  
The work of Bethel~\cite{bethel_survey_2008}, particularly her doctoral dissertation~\cite{bethel_robots_nodate}, is a seminal work in this field, as it explores the use of position, orientation, motion, colored light, and sound to add affective display to appearance-constrained robots used for search and rescue. There has also been some work applied in this space to drones, such as adding a head-like appendage to an aerial robot \cite{arroyo_daedalus:_2014}. 
This type of nonverbal communication has also been applied to a dog robot ~\cite{moshkina_human_2005} and learned over time by an agent given the feedback of a human ~\cite{shimokawa_acquiring_2001}. While many of these works focus on the actual display of the emotions and less on how they are generated, Novikova et al.~\cite{novikova_design_2014} models an emotional framework to generate a robot's emotions and then display them, largely through body language.

\subsubsection{Regulators and Adapters}
In this final category, we mostly find work which attempts to communicate some non-emotional state, which we consider to be adapters. A simple example of an adapter in human body language would be standing with slumped shoulders due to tiredness. While adapters can frequently wander into the realm of affective display, we consider a robot's display of its state to be an adapter-like communication. Works exist which merge the two, such as Knight et al~\cite{knight_expressive_2014}, which uses the motion of the robot to display the internal state and task state of the robot. A more straightforward example of an adapter, however, would be the work of Baraka et al.~\cite{baraka_mobile_2018}, which displays information such as intended path using an array of expressive lights around the robot's body. Regulators are not well explored in non-humanoid robots, but some works such as ~\cite{satake_how_2009} address the problem of how to initiate conversations.
\section{METHODOLOGY}
\begin{table*}[t]
	\centering
	\begin{tabular}{| l | l | c | c |}
        \hline
        \textbf{Meaning} &\textbf{ Kineme} & \textbf{Human Eqiv?} & \textbf{Meaning Type}\\ \hline
        \textit{Yes} & Head nod (pitch) & Yes & Response \\ \hline
        \textit{No} & Head shake (yaw) & Yes & Response\\ \hline
        \textit{Maybe} & Head bobble (roll) & Yes & Response \\ \hline
        \textit{Ascend} & Ascend, look back, continue & No & Spatial\\ \hline
        \textit{Descend} & Descend, look back, continue & No & Spatial\\ \hline
        \textit{Remain At Depth} & Circle and barrel roll slowly & No & Spatial\\ \hline
        \textit{Look At Me} & Roll heavily and erratically & No& Situation\\ \hline
        \textit{Danger Nearby} &Look around" then quick head shake & No & Situation \\ \hline
        \textit{Follow Me} & Beckon with head, then swim away& Yes & Spatial \\ \hline
        \textit{Malfunction} & Slowly roll over and pulse legs intermittently& No & State \\ \hline    
        \textit{Repeat Previous} & \enquote{Cock an ear} to the human& Yes & Response \\ \hline
        \textit{Object of Interest} & Orient toward object, look at human, proceed& Yes  & Spatial \\ \hline
        \textit{Battery Low} & Small, slow loop-de-loop & No & State \\ \hline
        \textit{Battery Full} & Large, fast loop-de-loop& No & State \\ \hline
        \textit{I'm Lost} & Look from side to side slowly as if confused & No & State \\ 
	    \hline
	\end{tabular}
    \caption{Kinemes with their associated meanings.}
    \label{tab:kin_meanings}
\end{table*}

In this section we introduce the design and implementation of our kineme communication system using robot motion and the light codes system to which we are comparing it.

\subsection{Kineme System}
\label{sec:kineme_system}
\subsubsection{Guiding Principles}
The development of meaningful motion is a somewhat out-of-scope problem for most computer scientists and engineers.  
It is most closely related to animation\cite{ribeiro_illusion_2012} in the way it much be approached, which is by considering how a motion is likely to effect the viewer.  
In this particular case of informative display for a non-humanoid underwater robot, there is little previous work, so it is important to develop guiding principles for design.

We applied the following concepts to develop our kinemes:

\begin{itemize}
	\item If a human analog for a gesture exists (such as nodding or shaking of the head), mimic it. 
    \item Exaggerate motions so that they are clearly visible from distance.
    \item Exploit any humanoid-looking design elements of the AUV; \textit{e.g.}, the front cameras of Aqua look somewhat like eyes, so \enquote{gazing} motions would likely work well.
\end{itemize} 

\subsubsection{Design Process}
With these concepts in mind, we selected a set of appropriate meanings for the kinemes.
We began by considering the types of information divers communicate with each other using hand signals as the basis for the information our robot should be able to communicate, identifying four primary categories: 
\begin{enumerate*}[label={\alph*)}]
	\item responses to queries,
	\item spatial information and commands,
	\item situational information and commands, and 
	\item state information.
\end{enumerate*}
With the possible meanings selected, we split them into those with obvious equivalent human gestures and those without.

The human equivalent group was developed by attempting to mimic the human gestures which existed. These kinemes were mostly in the Response category.
For these kinemes, Aqua's front was viewed as a face, with the cameras serving as eyes.  Then, by manipulating the motion of the whole robot, Aqua's \enquote{face} could be moved in a head-nodding motion for \textit{Yes}, a head-shaking motion for \textit{No}, etc.

The group without human equivalent gestures was more challenging to design motion for. For these kinemes, if they were spatially oriented, the general approach was to orient Aqua's \enquote{face} to that location, move towards it, (\textit{e.g.}, towards the surface to indicate \textit{Ascend}), look back at the human, and then continue towards the location. 
For kinemes in the situation and state categories, design started by identifying a relevant emotion, such as fear for \textit{Danger Nearby}, followed by developing a motion characteristic of that emotion. 
A complete list of kinemes along with their meanings can be found in Table \ref{tab:kin_meanings}.

\subsubsection{Implementation}
Implementation of kinemes was done using Epic Game's Unreal Engine$^{TM}$ to animate a 3D mesh of the Aqua AUV going through the selected motions. While physics simulation was not used to produce the kinemes, all motion was created by researchers familiar with the motion of the AUV in question and is achievable by the physical robot. The motion of Aqua's flippers was implemented as the forward swimming gait, regardless of the motion actually being executed. This is unlikely to have had an effect on participants, as none of them had experience with the Aqua robot, and almost all had never even seen it swim.

\begin{figure}
	\centering
	\begin{subfigure}[b]{0.49\textwidth}
    	\includegraphics[width=\textwidth]{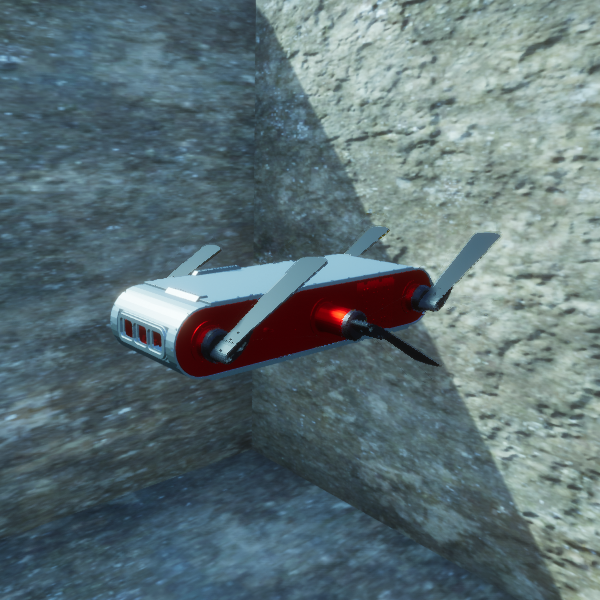}
		\caption{Unreal Engine$^{TM}$prototype.}
    	\label{fig:unreal}
	\end{subfigure}
    \hfill
    \begin{subfigure}[b]{0.49\textwidth}
    	\includegraphics[width=\textwidth]{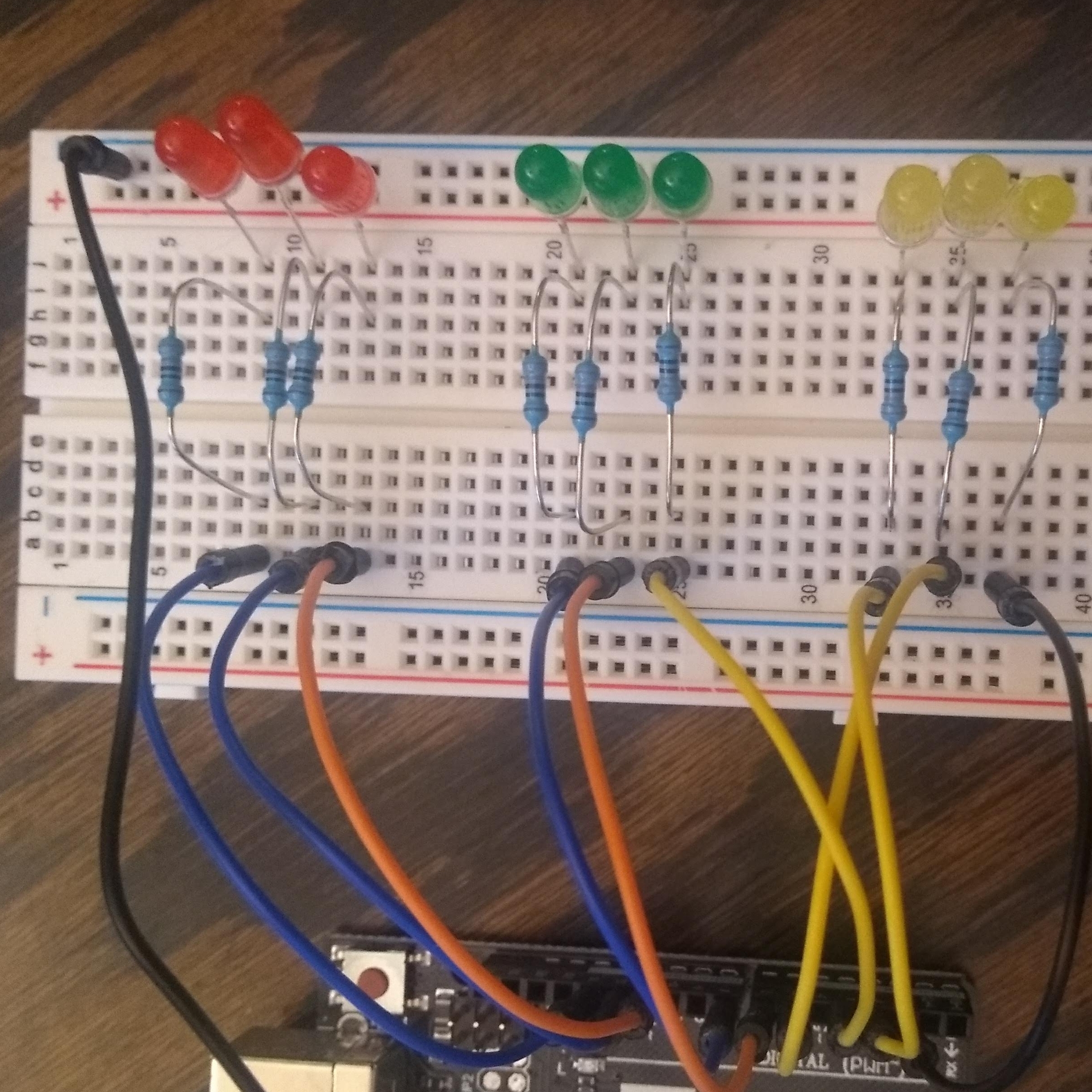}
		\caption{Arduino prototype.}
    	\label{fig:lights}
	\end{subfigure}
    \caption{Experimental platforms for Kineme and Light Codes.}
    \label{fig:platforms}
\end{figure}

\subsection{Lights System}

\subsubsection{Guiding Principles of Design}
While the light system was designed as a baseline to compare to the kineme system, care was taken to make the light system as robust as possible. 
To be used as a baseline system, the same meanings for the kineme system were selected. To guide the development of the light codes, a number of principles were used, based on human perception of color~\cite{sokolova_color_emotion_2015} and blink frequency~\cite{langmuir_light_aviation_1931}.
\begin{itemize}
\item Use natural color mappings (green = good, red = bad)
\item The faster the blinking, the more time sensitive the communication. 
\item For related information, share a portion of the light code (\textit{i.e.}, both battery info light codes have a single solid yellow light).
\end{itemize}

The list of light codes can also be found in Table \ref{tab:lights_meanings}.

\subsubsection{Implementation}
The light codes were chosen after the development of the kinemes and were implemented using an Arduino controlling 3 LEDs of each color.
Blink frequencies were selected subjectively, with a duration of five seconds on for solid lights, 1 Hz blinking for five seconds for slow blink rates, and 5 Hz blinking for four seconds for fast blink rates.

\begin{table*}[t]
	\centering
	\begin{tabular}{| l | l | c | c |}
        \hline
        \textbf{Meaning} &\textbf{Light Codes} & \textbf{Human Eqiv?} & \textbf{Meaning Type}\\ \hline
        \textit{Yes} & One solid green & Yes & Response \\ \hline
        \textit{No} & One solid red & Yes & Response\\ \hline
        \textit{Maybe} & One solid yellow & Yes & Response \\ \hline
        \textit{Ascend}& Two solid yellow, one blinking green & No & Spatial\\ \hline
        \textit{Descend} & Two solid yellow, one blinking green & No & Spatial\\ \hline
        \textit{Remain At Depth} & Two solid yellow, one blinking yellow & No & Spatial\\ \hline
        \textit{Look At Me} &Three quick flashing green & No & Situation\\ \hline
        \textit{Danger Nearby} & Three quick flashing red & No & Situation \\ \hline
        \textit{Follow Me} & Three blinking yellow & Yes & Spatial \\ \hline
        \textit{Malfunction} & Three solid red & No & State \\ \hline    
        \textit{Repeat Previous}& One blinking yellow light & Yes & Response \\ \hline
        \textit{Object of Interest}& Two solid green, one yellow blinking & Yes  & Spatial \\ \hline
        \textit{Battery Low} & One solid yellow, two blinking red & No & State \\ \hline
        \textit{Battery Full} & One solid yellow, two blinking green & No & State \\ \hline
        \textit{I'm Lost} & Three solid yellow & No & State \\ 
	    \hline
	\end{tabular}
    \caption{Light codes with their associated meanings.}
    \label{tab:lights_meanings}
\end{table*}
\section{EXPERIMENTAL SETUP}
To evaluate the kineme communication system as a method of robot-to-human communication,  a user study was conducted using the color light system as a baseline. This section describes the hypothesis being tested, the population, design of the study, and experimental methods.

\subsection{Hypotheses}
\label{sec:hypotheses}
The hypotheses we wish to test are simple: the accuracy of kinemes will be higher than that of light codes at all education levels (see Section~\ref{sec:experimental_methods}) and the operational accuracy (accuracy of answers with confidence $\geq$ $3$ on a scale of $1$ to $5$) of kinemes will be higher than light codes at all education levels.
We also hypothesize that the confidence of participants will be higher with kinemes than with light codes and that the time-to-answer for kinemes will be significantly longer than light codes at low education, but eventually fall to approximately the same time as the education of participants increases.

\subsection{Population}
The population for this study was 24 participants (16 male, 8 female), largely undergraduate and graduate students at the University of Minnesota. 
The mean age of participants was 22 (\textit{std\_dev}=3.3). 
To ensure that the participants were representative of non-expert users, participants were asked to rate their experience with robots on a scale of one to five ($\mu=1.54$, $\sigma=0.5$), as well as their experience with nonverbal communication ($\mu=1.33$, $\sigma=0.7$).

Participants were split into three groups, based on the amount of preparation for the communication task they would receive. 
We designate these groups EDU0, EDU1, and EDU2, each with 8 participants. 
Within each group, the order in which the systems were displayed was alternated, so that half would see the kinemes first and half would see the light codes first.  

\subsection{Experimental Methods}
\label{sec:experimental_methods}
After being enrolled in the study and providing basic demographic information from a survey, each participant was provided with the same basic information about the problem, namely that the purpose of the study was to develop underwater communication for robots. Next, participants received the appropriate level of education:

\begin{itemize}
\item EDU0: Participants were told the communication vector (motion, lights).
\item EDU1: Participants were told the communication vector as well as the list of possible meanings.
\item EDU2: Participants were told the communication vector and shown videos of each kineme and light code while being told the meaning.
\end{itemize}

Education was offered directly before testing each system. Once a participant was oriented and educated to a system, they were shown videos of the kinemes or light codes in a random order.  
The random order of the videos limits the order dependencies of the kinemes and light codes and produces independent measurements of each kineme.

For each video, three pieces of data were recorded: the user's understanding of what was communicated, the time taken from the start of the video to the start of the participant answering, and their confidence in their answer from $1$ to $5$, with five being positive. After-the-fact correctness for each answer was assigned according to some simple heuristics by an expert user.

\subsection{Additional Modality} 
A third communication system employing messages display on an LCD screen was also tested in the study. The results from the LCD are not reported here, because it was not a realistic test of an LCD for underwater use. The purpose of the LCD in this study was simply to act as a control by which reaction times and confidence distributions for participants could be considered, as well as acting as a filter for participants who completely misunderstood the purpose of the experiment. 
\section{RESULTS}

\subsection{Comparison Criteria and Methods}
The kineme and light system are compared on the basis of the following criteria:

\begin{itemize}
\item \textbf{Accuracy} -- The accuracy of a participant's understanding of a kineme or light code, rated from 0 to 10 in order of increasing accuracy. 
\item \textbf{Confidence} -- The confidence a participant has in their understanding of a kineme or light code, rated from 1 to 5, in order of increasing confidence.
\item \textbf{Operational Accuracy} -- The same metric as accuracy, but only taking answers rated at a confidence level of 3 or higher, representing the answers that participants would be likely to act on.  
\item \textbf{Time To Answer} -- The time it takes a participant to give the meaning of a kineme or light code, measured in seconds from the beginning of the signal to the beginning of their answer.
\end{itemize}

We use the Mann-Whitney test \cite{HettmanspergerThomasP.1939-1998Rnsm} to evaluate the hypotheses we set out in Section~\ref{sec:hypotheses}. We also use the Mann-Whitney test to validate that there is no statistically significant improvement in accuracy regardless of which system was shown to participants first and that there is no statistically significant difference between the accuracies of male and female participants.

The Mann-Whitney test is ideal for measuring the statistical effects of using the different systems in our trials, as it does not require normally-distributed data, while providing hypothesis testing capabilities. For each Mann-Whitney test, we report the p-value $p$, which is the probability under the null hypothesis of obtaining a result equal to or more extreme than what was observed. We also report the z-statistic $z$, which is used to calculate the approximate p-value.  
$z$ is defined as

$$ z = \frac{W - E(W)}{\sqrt[]{V(W)}}$$

where W is the Wilcoxon rank sum. Unless otherwise noted, all hypothesis tests are conducted at significance level $\alpha = 0.005$.
\begin{figure}[t!]
\centering
     \begin{subfigure}[t]{0.49\textwidth}
     \centering
		\includegraphics[width=\textwidth]{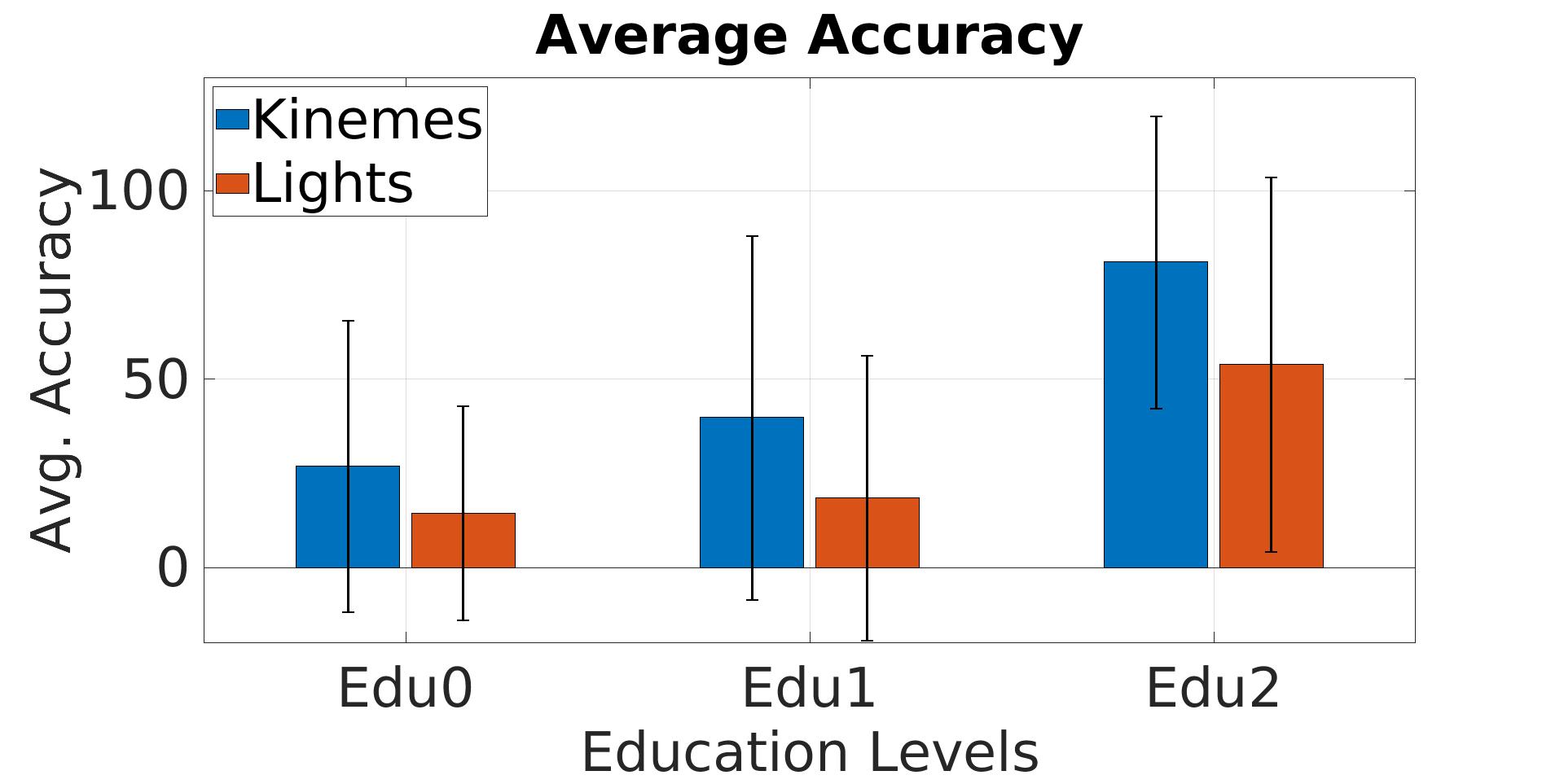}
    	\caption{Average Accuracy per education level}
        \label{fig:avg_acc}
	\end{subfigure}
    \hfill
	\begin{subfigure}[t]{0.49\textwidth}
    \centering
		\includegraphics[width=\textwidth]{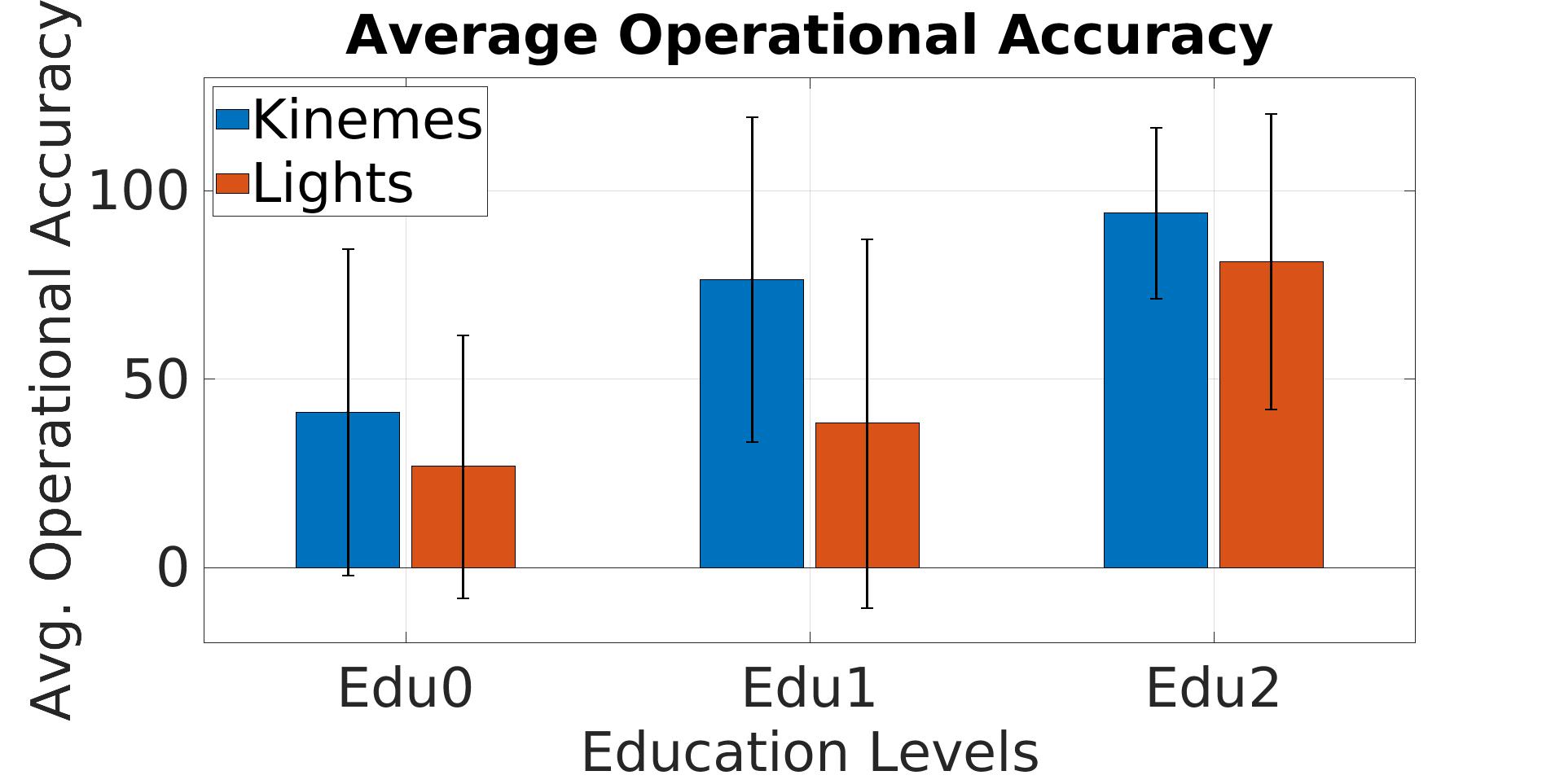}
    	\caption{Average Operational Accuracy per education level.}
        \label{fig:avg_op}
	\end{subfigure}
	    \caption{Average Accuracy and Operational Accuracy per education level.}
    \label{fig:accuracies}
\end{figure}

\subsection{Statistical Results}
\subsubsection{Accuracy and Operational Accuracy Between Education Levels}
   
We compare kineme and light code accuracies at each education level, using a right-tailed Mann-Whitney test with this hypothesis: 
    \begin{gather*}
    	H_{0} = \text{Kineme accuracy does not have a higher median.}\\
      	H_{a} = \text{Kineme accuracy has a higher median than lights.} 
	\end{gather*}
        
When testing accuracy, we find statistically significant increases in accuracy when comparing kinemes to light codes for EDU0 ($p=0.0113, z=2.282$), EDU1 ($p=0.0009, z=3.114$), and EDU2 ($p=0.000009,4.27$). 
For operational accuracy, we again find statistically significant increases for EDU0 ($p=0.029,z=1.890$), EDU1 ($p=0.007,z=2.418$), and EDU2 ($p=0.0006,z=3.221$). 
In all of these cases, we can reject the null hypothesis in favor of the alternative. 
We can also see this visually in the plots of these statistics in Figures \ref{fig:avg_acc} and \ref{fig:avg_op}.

\subsubsection{Confidence and Time-To-Answer Between Education Levels}
\begin{figure}[t!]
\centering
     \begin{subfigure}[t]{0.49\textwidth}
     \centering
		\includegraphics[width=\textwidth]{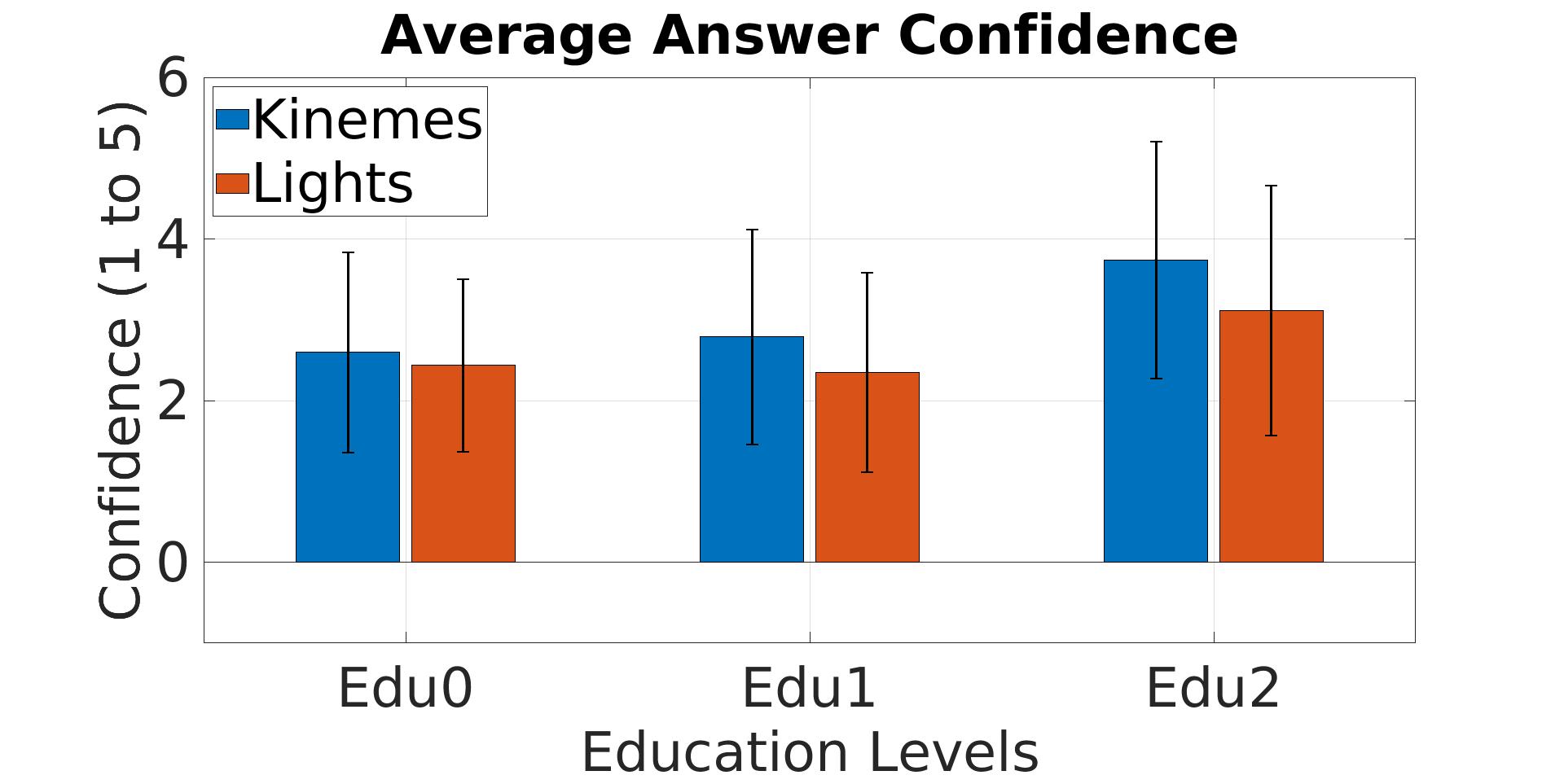}
    	\caption{Average Confidence per education level.}
        \label{fig:avg_conf}
	\end{subfigure}
    \hfill
	\begin{subfigure}[t]{0.49\textwidth}
    \centering
		\includegraphics[width=\textwidth]{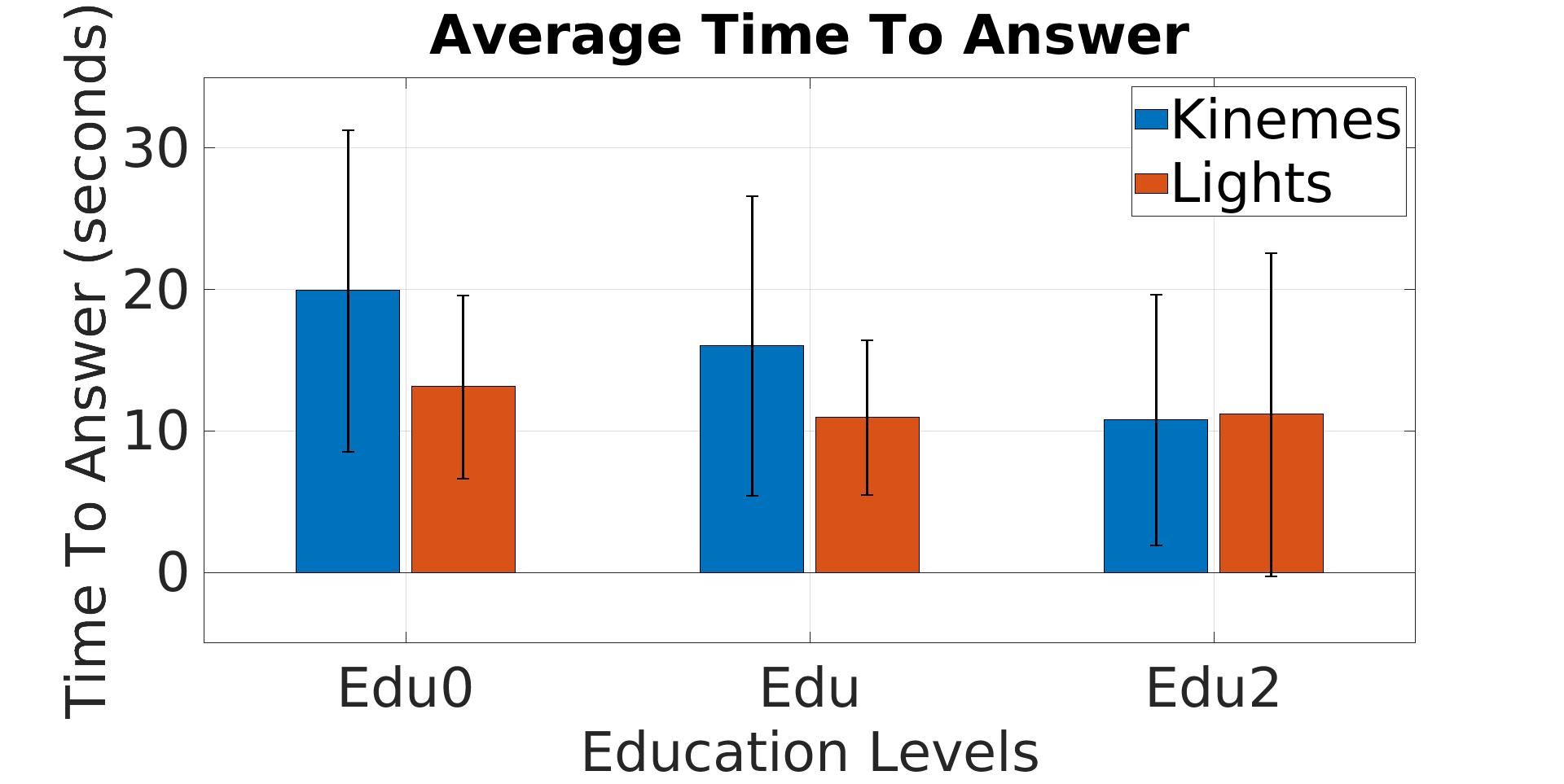}
    	\caption{Average Time To Answer per education level.}
        \label{fig:avg_time}
	\end{subfigure}
	    \caption{Average Confidence and Time To Answer per education level.}
    \label{fig:conf_time}
\end{figure}

We also test the median of confidence participants reported in their answers, and here we find that while there is no statistically significant increase in confidence in kinemes vs light codes at EDU0 ($p=0.156,z=1.01$), there is a statistically significant increase present at EDU1 ($p=0.006,z=2.496$) and EDU2 ($p=0.0004,z=3.297$). 
Finally, when considering the time to answer, we must slightly reformulate our test to be a left tailed test, testing the null hypothesis that the median time to answer for light codes is not lower than for kinemes, with the alternative being that the median for lights is lower. 
Here, we show that there is a statistically significant reduction in time when comparing lights to kinemes for EDU0 ($p=0.0000002,z=-5.031$) and EDU1 ($p=0.0003,z=-3.39$), but at EDU2 ($p=0.4074,z=-0.828$) there is no such statistically significant reduction. 
This indicates that the difference between the time to answer falls at each education level, with the difference eventually dropping below statistical significance. Again, we can see both of these trends in Figures \ref{fig:avg_conf} and \ref{fig:avg_time}.

\subsubsection{Comparison Between Specific Kinemes and Codes}
\begin{figure}
	\centering
	\begin{subfigure}[b]{\textwidth}
      \includegraphics[width=\textwidth]{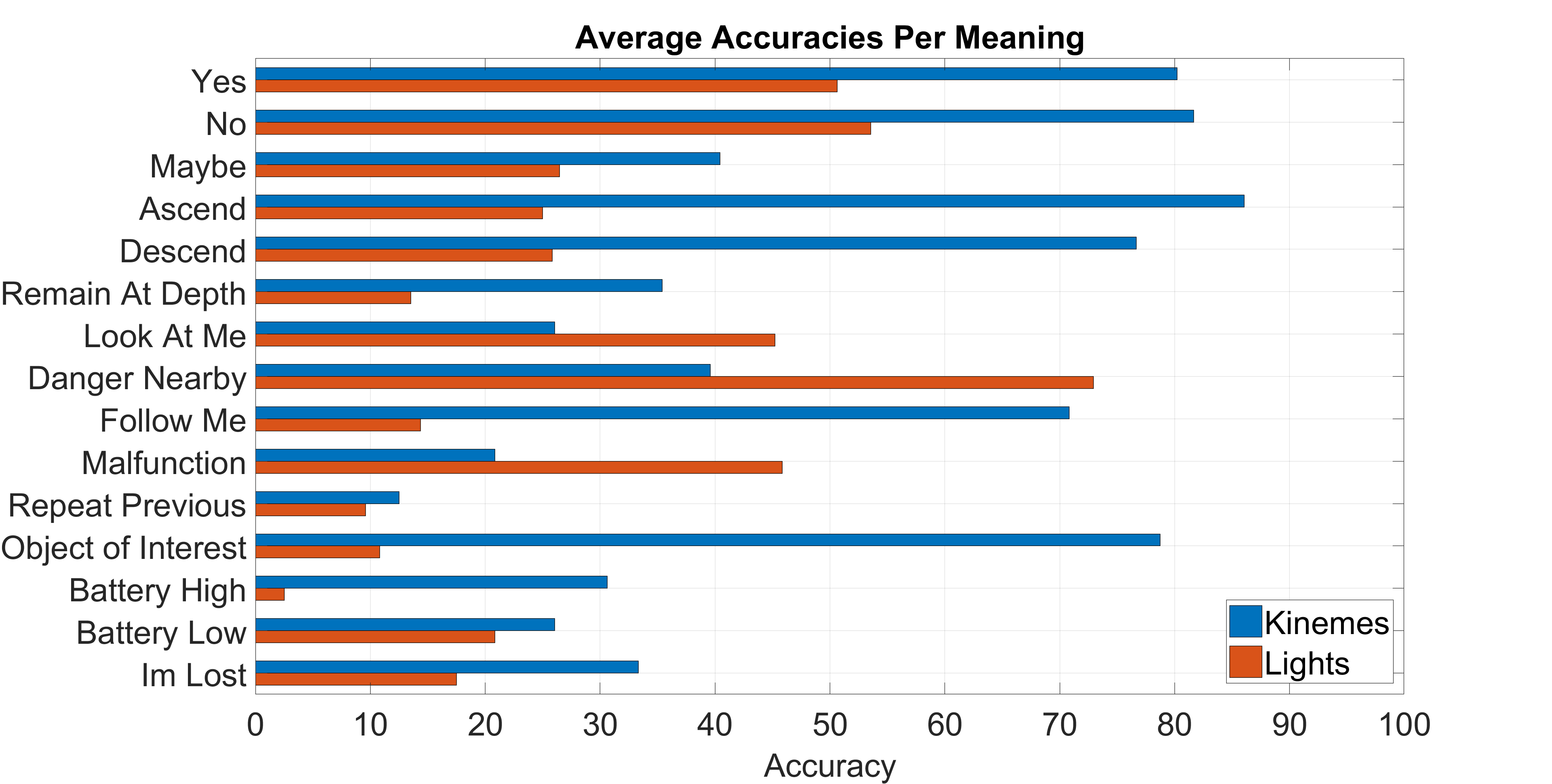}
      \caption{Average Accuracy per meaning.}
      \label{fig:per_acc}
	\end{subfigure}
    
    \begin{subfigure}[b]{\textwidth}
      \includegraphics[width=\textwidth]{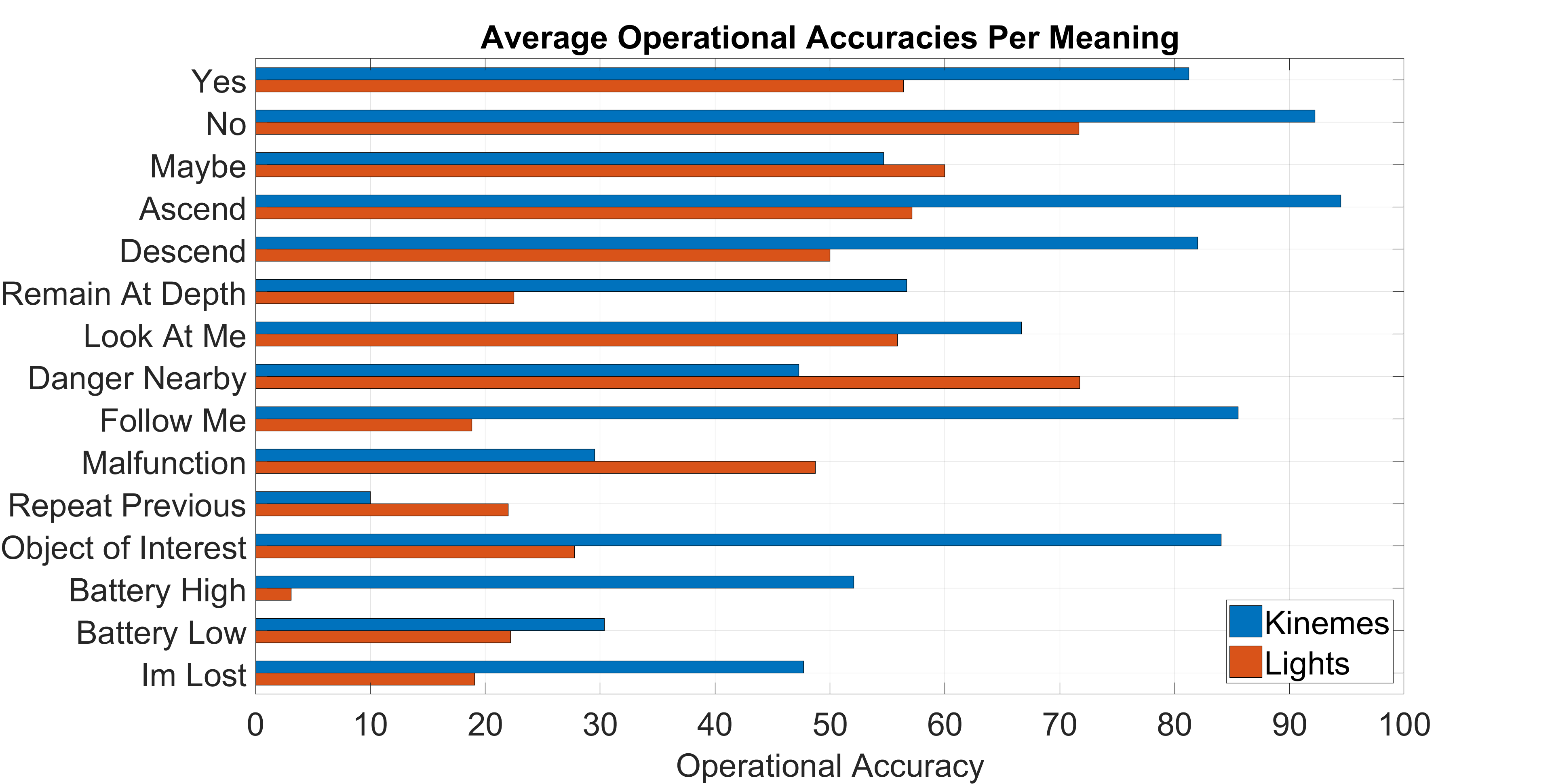}
      \caption{Average Operational Accuracy per meaning.}
      \label{fig:per_op}
	\end{subfigure}
    \caption{Average Accuracy and Operational Accuracy per meaning.}
    \label{fig:platforms}
    \vspace{-5mm}
\end{figure}

    For kineme-by-kineme comparison, we direct the reader to Figures \ref{fig:per_acc} and \ref{fig:per_op}.  
We can see a particularly high accuracy for those kinemes in the spatial category, paired with low light code accuracy for those same meanings.  
Conversely, we see that situation concepts such as \textit{Danger Nearby} and \textit{Malfunction} work much better with flashing lights, likely due to a lifetime of being taught to watch out for flashing red lights. 
Operational accuracy figures are much closer between light codes and kinemes, but whether the kineme or light codes are the most accurate system does not change between accuracy and operational accuracy.

\subsubsection{Internal Validation For Order and Gender}
    We validate these results by checking for a statistically significant bias based on system order. We find no statistically significant difference between the accuracy of kinemes when shown first or when shown second ($p=0.449,z=-0.756$), nor do we find a statistically significant change in the accuracy of the light codes ($p=0.748, z=-0.320$). 

Additionally, we find no bias in accuracy for male vs female participants, both with the whole system ($p=0.168, z=1.375$), with kinemes ($p=0.150, z=1.436$) and light codes($p=0.808, z=0.242$) considered separately. 

\subsection{Opinion-Based Results}
	\subsubsection{Participant Opinions}
 	In their exit survey, participants were asked to rate the kineme and lights systems on a scale of 1 to 10 for several metrics. Participants rated kinemes easier to understand ($\mu=5.6 ,\sigma=2.2 $) than lights ($\mu=3.5 ,\sigma=3.3 $). They also considered kinemes easier to learn ($\mu= 7,\sigma=2.1 $) than lights ($\mu=5.5 ,\sigma=3.5 $).
	When asked, 71.4\% of participants also preferred the light system overall, 66.7\% felt it would be most effective from a significant distance, and when asked where kinemes, light codes, or an LCD would be best for underwater communication system, 45.6\% of participants preferred the kineme system, compared to 37.5\% for lights and 16.7\% for the LCD. 
    

\section{CONCLUSION}
In this paper, we proposed a unique motion-based communication for underwater robots, which we call kinemes, and implemented a version of these kinemes in Unreal Engine $^{TM}$ for testing. 
We evaluated the use of kinemes versus the use of colored light codes and found statistically significant superiorities in accuracy and operational accuracy, while remaining within an acceptable speed of recognition. 
Additionally, in our study, users preferred the kineme system over the light code system, and even over an LCD screen, especially when considering use at a distance or underwater.

We have also found that certain concepts related to $3$ dimensional space are especially easy to communicate through motion via perceived gaze directions, as are concepts with a direct human analogue by mimicking that human analogue. 
Other concepts, such as reporting danger, might be better expressed through some other vector. 
In the design of a motion-based communication system, it is recommended that designers consider any possible human analogues to the meaning they want to convey and exploit human-like features in their robot's design. 
In the future, we plan to extend this concept to other 6-DOF systems and implement kinemes on the physical Aqua robot, further validating our findings by running studies involving fully closed-loop interaction tests and more participants. Furthermore, we plan to integrate light and sound alongside motion to create a communication system which uses all possible nonverbal communication vectors to effectively communicate information  to human collaborators in underwater environments.  
\section*{ACKNOLWEDGMENTS}
The authors gratefully acknowledge the support of the MnDRIVE initiative on this research; Hannah Dubois, Marc Ho, and Jungseok Hong for their help in conducting studies; Md Jahidul Islam for his advice; The Mobile Robotics Lab at McGill University, Independent Robotics, and particularly Ian Karp for the use of their Unreal Engine robot model.

\bibliographystyle{abbrv}
\bibliography{RCVM}

\end{document}